\documentclass[smallabstract,smallcaptions]{dccpaper}
\pdfoutput=1
\usepackage{epsfig}
\usepackage{amsmath}
\usepackage{amssymb}
\usepackage{color}
\usepackage{url}
\usepackage{subcaption}
\newlength{\figurewidth}
\newlength{\smallfigurewidth}

\begin{document}

\title
{\large
\textbf{An Empirical Analysis of Recurrent Learning Algorithms In Neural Lossy Image Compression Systems }
}

\author{Ankur Mali\textsuperscript{$\star$},
Alexander G. Ororbia\textsuperscript{\textdagger}, Dan Kifer \textsuperscript{$\star$},
C. Lee Giles\textsuperscript{$\star$}\\ [0.5em]
\textsuperscript{$\star$\ The Pennsylvania State University, University Park, PA, 16802, USA}\\
\textsuperscript{\textdagger Rochester Institute of Technology, Rochester, NY, 14623, USA}\\
}

\maketitle
\thispagestyle{empty}

\begin{abstract}
   Recent advances in deep learning have resulted in image compression algorithms that outperform JPEG and JPEG 2000 on the standard Kodak benchmark. However, they are slow to train (due to backprop-through-time) and, to the best of our knowledge, have not been systematically evaluated on a large variety of datasets. In this paper, we perform the first large scale comparison of recent state-of-the-art hybrid neural compression algorithms, while exploring the effects alternative training strategies (when applicable). The hybrid recurrent neural decoder is a former state-of-the-art model (recently overtaken by a Google model) that can be trained using backprop-through-time (BPTT) or with alternative algorithms like sparse attentive backtracking (SAB), unbiased online recurrent optimization (UORO), and real time recurrent learning (RTRL). We compare these training alternatives along with the Google models (GOOG and E2E) on 6 benchmark datasets. Surprisingly, we found that the model trained with SAB performs the better (outperforming even BPTT), resulting in faster convergence and better peak signal-to-noise ratio.

\end{abstract}

\section{Introduction}

Recently, deep neural networks (DNNs) have been shown to yield excellent performance across many domains including computer vision, speech recognition, and natural language processing. 
%
However, in areas such as image compression, deep learning-based approaches struggle with large training times, which is problematic since generalizability to different image datasets has not yet been established (to the best of our knowledge, there has been no large scale evaluation of the state-of-the art). 
Thus, in this work, we conduct a large-scale empirical comparison of a variety of training algorithms for two state-of-the-art neural compression architectures: the hybrid neural decoder  \cite{orobiadcc, malidcc} and a recently proposed Google model \cite{toderici2016full,balle2016end}. The training algorithms include backprop-through-time (BPTT) as well as BPTT alternatives. 

BPTT is the go-to algorithm for training recurrent neural architectures. Nonetheless, despite its popularity, it can be slow and unstable (vanishing/exploding gradients). Alternatives such as RTRL \cite{williams1989experimental}, UORO \cite{tallec2017unbiased}, and SAB \cite{ke2017sparse} have been proposed to address various drawbacks of BPTT.
However, these alternative training algorithms have their own limitations. For example, it is difficult to use them to train extremely large models with convolutions, such as the powerful Google models GOOG \cite{toderici2016full} and E2E \cite{balle2016end}. 
Thus, an important question is whether one should focus on designing architectures that are compatible with these backprop alternatives -- does alternative training provide benefits that compensate for a decrease in model flexibility?

The hybrid neural decoder  (HBD) \cite{orobiadcc, malidcc} is a former state-of-the-art model. In its original conception, HBD was trained with backprop-through-time, but we note that it is also easily trainable with the other three main alternatives -- UORO, RTRL, and SAB. Thus, we compare it to GOOG and E2E trained with BPTT in this study.
While previous studies were generally limited to the Kodak dataset, our study uses $6$ benchmarks, allowing us to examine how consistent these methods are in terms of convergence speed and generalization error. Furthermore, this allows us to examine how transferable the models are from one dataset to another.
In addition to this comparison, we propose an extension to the hybrid neural decoder that makes it compatible with both offline and online learning algorithms.

From our results, we observe that SAB takes 170 epochs on average to converge which is more compared with other approaches including BPTT, but requires least iterative refinement steps and obtains better compression ratio. SAB on average is 0.28dB better than BPTT. On the other hand RTRL converges on average at 120 epochs, compared with UORO which takes 160 and BPTT which takes 162. Despite reaching similar performance during training, UORO and RTRL struggle to match BPTT performance on test, which also hampers transfer performance on unseen data.


\subsection{Related Work}

Widely used lossy image compression methods such as JPEG and JPEG2000 (JP2) employ a combination of fixed transformations using entropy-based encodings to achieve better compression \cite{takamura1994coding}. This is suitable for real-time processing when memory and computational efficiency is needed. Recently, DNN approaches have outperformed these traditional methods for image compression. However, most of this work has focused on designing end-to-end systems that reconstruct images in a two-dimensional space \cite{oord2016pixel} using architectural building-block models such as auto-encoders, convolutional networks, and recurrent networks.
One bit of early work crafted a framework based on variational autoencoders to achieve better compression \cite{gregor2016conceptualcompression}.

Early work which outperformed classical techniques (at low bit rates), without harming perceptual quality, set the widely-adopted practice for using deep artificial neural networks (ANNs) in compression \cite{toderici2015,toderici2016full,johnston2018improved}.
Other, later methods \cite{Theis2017,rippel2017real,balle2018variational,agustsson2017soft} that followed focused on using convolutional networks or generative adversarial networks (GANs). Recent work has used spatial-temporal energy compaction \cite{cheng2019learning}, other energy compaction-based techniques \cite{chen19}, and filter-bank based convolution networks \cite{dcc19}. The majority of these end-to-end solutions have been designed to extract better latent representations and to eliminate redundancies in compression.

A simpler approach to the compression approach takes into account redundancy at the decoder side of the system and attempts to iteratively decompress using a hybrid recurrent decoder \cite{orobiadcc, malidcc}. Similarly, a standard encoder can be replaced with another DNN to enhance model representation and decode information using a standard decoder \cite{strumpler2020learning}. Prior efforts have shown the limitations of backprop based approaches on standard computer vision tasks \cite{lee2015difference,nokland2016direct,ororbia2019biologically} as well as natural language processing benchmarks \cite{ororbia2018continual}. This line of backprop-alternative work has motivated us to test the effect of promising backprop-alternative learning approaches on large scale datasets in a more challenging domain, i.e image compression.  Note that all neural-based compression systems that largely composed of auto-encoders and convolutional networks, mentioned above, are trained using traditional BP and systems composed of recurrent networks are trained via BPTT \cite{werbos1988generalization,werbos1990backpropagation} . In this work, we focus on alternatives to BPTT for training recurrent networks \cite{ke2017sparse,williams1989experimental,tallec2017unbiased} and analyze the effect that these algorithms have when crafting a neural-based compression system. We next describe our neural decoding system based on \cite{orobiadcc, malidcc}.

\section{Hybrid Nonlinear Estimator for Iterative Decoding}
\label{iterative_refinement}

\noindent
\textbf{Iterative Refinement:} This procedure can be seen as locally decoding data process aimed at improving the memory retention ability of recurrent neural networks (RNNs) \cite{orobiadcc,mali2019neural, malidcc}. The neural decoder used with this process essentially reconstructs images from a compressed representation and iterative refinement formulates compression as a multi-step reconstruction problem over a finite number of passes, $K$. 
%
Consider
a 2D image $\mathbf{I}$ and decompose it into a set of $P$ image patches, or $\mathbf{I} = \{ \mathbf{p}^1,\cdots,\mathbf{p}^j,\cdots,\mathbf{p}^P \}$ (i.e non-overlapping for JPEG, overlapping for JP2). By assuming a column-major orientation, each input patch has dimension $\mathbf{p}^j \in \mathcal{R}^{d^2 \times 1}$ . Then each patch $\mathbf{p}^j$ would have a corresponding quantized symbol representation $\mathbf{q}^j$ of dimension $\mathbf{q}^j \in \mathcal{R}^{d^2 \times 1}$. 
%

The neural decoder is defined by parameters $\Theta = \{\Theta_s, \Theta_t, \Theta_d\}$ taking in $N$ neighboring patches as input. The estimator's form requires $3$ key functions \cite{orobiadcc}: 
\begin{itemize}
\item $\mathbf{e} = e(\mathbf{q}^1, \cdots, \mathbf{q}^N ; \Theta)$, is an embedding of the quantized symbols for N neighboring patches. It is typically called an \emph{transformation function}.
\item $\mathbf{s}_k = s(\mathbf{e}, \mathbf{s}_{k-1} ; \Theta)$, is the recurrent state function that combines the  embedding with the previous state (like an RNN).
\item $\widetilde{\mathbf{p}}^j_k = d(\mathbf{s}_k ; \Theta)$, a reconstruction function that predicts a target patch at step $k$.
\end{itemize}
Both the transformation function $\mathbf{e} = e(\mathbf{q}^1, \cdots, \mathbf{q}^N ; \Theta)$ and the reconstruction function $\mathbf{s}_k = s(\mathbf{e}, \mathbf{s}_{k-1} ; \Theta)$ can be parametrized by multilayer perceptrons (MLPs),
using parameters $\Theta_t = \{W_1, \cdots, W_N\}$ and $\Theta_d = \{U, \mathbf{c}\}$, respectively:
\begin{equation}
\mathbf{e} = \phi_e(W_1 \mathbf{q}_1 + \cdots + W_N \mathbf{q}_N) \mbox{, and, } 
\widetilde{\mathbf{p}}_k = \phi_d(U \mathbf{s}_k + \mathbf{c})
\nonumber 
\end{equation}
where $\phi_e(v) = v$ and $\phi_d(v) = v$.

\subsection{State Function Forms}
\label{state_fun}
We experimented with a variety of gated recurrent architectures and unify the majority of recurrent architectures under the Differential State Framework (DSF) \cite{ororbia2017diff}.
We experimented with popular recurrent structures including Long Short Term Memory (LSTM) model \cite{hochreiter1997long}, Gated Recurrent Units (GRU) \cite{chung2014empirical}, and Delta-RNN (Delta-RNN or $\Delta$-RNN). We compared these RNN-based models to a static mapping function learned by a stateless MLP.

The $\Delta$-RNN state function (parameters $\Theta_s = \{V, \mathbf{b}, \mathbf{b}_r, \alpha, \beta_1, \beta_2 \}$) is defined as:
\begin{align}
\mathbf{d}^1_k &= \alpha \otimes V \mathbf{s}_{k-1} \otimes \mathbf{e}, \quad 
\mathbf{d}^2_k = \beta_1 \otimes V \mathbf{s}_{k-1} + \beta_2 \otimes \mathbf{e} \\
\widetilde{\mathbf{s}}_k &= \phi_s(\mathbf{d}^1_k + \mathbf{d}^2_k + \mathbf{b}) \\
\mathbf{s}_k &=  \Phi( (1 - \mathbf{r}) \otimes \widetilde{\mathbf{s}}_k + \mathbf{r} \otimes \mathbf{s}_{k-1} ), \ \mbox{where,} \ \mathbf{r} =  \sigma(\mathbf{e} + \mathbf{b}_r) \label{general_second_order},
\end{align}
where $\Phi(v) = \phi_s(v) = tanh(v) = (e^{(2v)} - 1) / (e^{(2v)} + 1)$ and $\otimes$ denotes the Hadamard product.
\paragraph{The RNN State Function}
\label{rnn}
We can easily parameterize the above three functions in a classical Elman-style RNN, which requires defining the state parameters to be $\Theta_s = \{V, \mathbf{b}\}$. The state function of an Elman-RNN is quite simple:
\begin{align}
\mathbf{s}_t &= \phi_s(\mathbf{e} + V \mathbf{s}_{t-1} + \mathbf{b})
\label{state:rnn}
\end{align}
where the overall state is a linear combination of the transformation function's output and affine transformation of the filtration ($V \mathbf{s}_{t-1} + \mathbf{b}$). The post-activation function $\phi_s$ can be any differentiable element-wise function, such as the logistic sigmoid $\phi_s(v) = \sigma(v) = 1/(1 + e^{-v})$, the hyperbolic tangent $\phi(v) = tanh(v) = (e^{(2v)} - 1) / (e^{(2v)} + 1)$, or the linear rectifier $\phi_s(v) = relu(v) = max(0, v)$.

\paragraph{LSTM State Function}
Because of its performance the LSTM \cite{hochreiter1997long} is one of the most commonly used gated neural model when modeling sequential data. The original motivation behind the LSTM was to implement the ``constant error carousal'' in order to mitigate the problem of vanishing gradients. This means that long-term memory can be explicitly represented with a separate cell state $\mathbf{c}_t$. 

The LSTM state function (without any extensions, such as ``peephole'' connections) is implemented using the following equations:
\begin{align}
\mathbf{f}_t &= \sigma(\mathbf{e} + V_f \mathbf{s}_{t-1} + \mathbf{b}_f), \quad
\mathbf{i}_t = \sigma(\mathbf{e} + V_i \mathbf{s}_{t-1} + \mathbf{b}_i) \\
\widetilde{\mathbf{c}}_t &= tanh(\mathbf{e} + V_c \mathbf{s}_{t-1} + \mathbf{b}_c), \quad
\mathbf{c}_t = \mathbf{f}_t \otimes \mathbf{c}_t + \mathbf{i}_t \otimes \widetilde{\mathbf{c}}_t \\
\mathbf{o}_t &= \sigma(\mathbf{e} + V_o \mathbf{s}_{t-1} + \mathbf{b}_o) , \quad
\mathbf{s}_t = \mathbf{o}_t \otimes \phi_s(\mathbf{c}_t)
\label{state:lstm}
\end{align}
where we depict the sharing of the transformation function's output across the forget ($\mathbf{f}_t$), input ($\mathbf{i}_t$), cell-state proposal ($\widetilde{\mathbf{c}}_t$), and output ($\mathbf{o}_t$) gates. However, if weight-tying is not used for the input-patch to hidden weights, we would assign specific matrices accordingly per gate (much as the recurrent weight matrices are assigned per gate). Note that $\sigma(v) = 1/(1 + e^{-v})$ and $tanh(v) = (e^{(2v)} - 1) / (e^{(2v)} + 1)$.

\paragraph{GRU State Function}
The Gated Recurrent Unit (GRU; \cite{chung2014empirical}) can be viewed as an attempt to simplify the LSTM. Among the changes made, the model fuses the LSTM input and forgets gates into a single gate, and merges the cell state and hidden state back together.
The state function based on the GRU can be calculated using the following equations:
\begin{align}
\mathbf{z}_t &= \sigma(\mathbf{e} + V_z \mathbf{s}_{t-1} + \mathbf{b}_z), \quad \mathbf{r}_t = \sigma(\mathbf{e} + V_r \mathbf{s}_{t-1} + \mathbf{b}_r) \\
\widetilde{\mathbf{s}}_t &= \phi_s(\mathbf{e} + V_s (\mathbf{r}_t \otimes \mathbf{s}_{t-1}) + \mathbf{b}_s), \
\mathbf{s}_t = \mathbf{z}_t \otimes \widetilde{\mathbf{s}}_t + (1 - \mathbf{z}_t) \otimes \mathbf{s}_{t-1}.
\label{state:gru}
\end{align}
Similar to the LSTM function described, the transformation function is depicted as shared across all internal gating functions. However, this is not always the case and separate gate-specific input-to-hidden parameters can be used. Note that $\phi_s(v) = tanh(v)$.

\paragraph{MLP Stateless Function}
\label{mlp}
The simplest stateless function can be parametrized by a single hidden layer multilayer perceptron (MLP) and only exploits non-causal context. The model is simply represented as follows:
\begin{align}
\mathbf{s}_t &= \phi_s(\mathbf{e} + \mathbf{b}).
\label{state:mlp}
\end{align}


\noindent
\textbf{Learning the Neural Iterative Decoder:} We explicitly unroll our estimator over the length of iterative refinement steps $K$ to create a mini-batch of length $K$ arrays of $B$ matrices, i.e., 3D tensors, in order to use BP(TT) to learn $\Theta$ (same for SAB). However, for online learning approaches (UORO and RTRL), we do not unroll over steps and directly compute gradients at each time step, forward propagating the gradients.   
Our objective will be to optimize distortion $D$, since we are crafting an estimator only for the act of decoding. Note, the estimator must learn to deal with variable bit-rates (as dictated by training samples).  
The mean bit-rate of our training dataset was $R_\mu = 0.525$ with variance $R_{\sigma^2} = 0.671$. 

At training time, we optimize decoder parameters with respect to a multi-objective loss over $K$-step reconstruction episodes for mini-batches of $B$ target patches $\mathbf{p}^j$ (channel input) operating over a set of decoder reconstructions $\widehat{\mathbf{p}}^j = \{ \widetilde{\mathbf{p}}^j_k, \cdots, \widetilde{\mathbf{p}}^j_K \}$ (channel outputs). The loss is defined as a convex combination of mean squared error (MSE) and mean absolute error (MAE) as follows:
\begin{align}
\mathcal{D}(\mathbf{p}^j,\widehat{\mathbf{p}}^j) = (1 - \alpha) \mathcal{D}_{MAE}(\mathbf{p}^j,\widehat{\mathbf{p}}^j) + \alpha \mathcal{D}_{MSE}(\mathbf{p}^j,\widehat{\mathbf{p}}^j) \mbox{.} \label{cost_function}
\end{align}
$\alpha$ (set to $0.235$) controls the tradeoff between the distortion terms.
MSE \& MAE are used in \cite{orobiadcc} and MSE was used \cite{malidcc} -- we follow the framework of these prior efforts.
\textbf{Real-Time Recurrent Learning:} 
Real-time recurrent learning (RTRL) \cite{williams1989experimental} is an online learning procedure for training recurrent neural networks (RNNs). Unlike BPTT , RTRL does forward propogation and does not suffer from a deep credit assignment problem. RTRL optimizes weights, denoted as $\Theta$, by minimizing a total loss (for stateful models) defined as follows:
\begin{align}
\mathbf{z}_{t+1} = F_{state}(\mathbf{x}_{t+1},\mathbf{z}_{t},\Theta) \mbox{.} \label{rtrl:eqn1}
\end{align}
where we note that $\mathbf{x}_{t+1}$ represents any possible vectorized input but, in this study, will contain the image patch(es) as described in the last section ($t$ in this case could be replace with $k$ in the context of iterative refinement). This will be the same for all learning procedures subsequently described.

One key advantage of RTRL is that it computes the derivative of the states and the outputs in its forward computation, thus eliminating the unfolding of the graph.
For next step prediction, the loss $L$ to optimize using RTRL, is: 
\begin{multline}
\frac { \partial L _ { t + 1 } } { \partial \Theta } = \frac { \partial L_{t + 1}(\mathbf{y}_{t + 1}, \mathbf{y}_{t + 1}^{*} )} {\partial \mathbf{y}}  \otimes \bigg( \frac{\partial F_{\text{out}}(\mathbf{x}_{t + 1}, \mathbf{z}_{t}, \Theta)} {\partial \mathbf{z}_t} \frac{ \partial \mathbf{z}_{t} } {\partial \Theta} + \frac {\partial F_ {\text{out}}(\mathbf{x}_{t + 1} , \mathbf{z}_{t} , \Theta)}{\partial \Theta} \bigg) \mbox{.} \label{rtrl:eqn2}
\end{multline}
If we differentiate Equation \ref{rtrl:eqn1} with respect to $\Theta$, we obtain:
\begin{align}
\frac{\partial \mathbf{z}_t+1}{\partial \Theta} = \frac{\partial F_{\text {state}}(\mathbf{x}_{t+1}, \mathbf{z}_{t}, \Theta)}{\partial \Theta} + \frac{\partial F_{\text {state}}(\mathbf{x}_{t+1}, \mathbf{z}_{t}, \Theta)}{\partial \mathbf{z}_t} \otimes \frac{\partial \mathbf{z}_t}{\partial \Theta} \label{rtrl:eqn3}
\end{align}
where for step we compute $\frac{\partial \mathbf{z}_t}{\partial \Theta}$ based on $\frac{\partial \mathbf{z}_t-1}{\partial \Theta}$ and use these values to compute $\frac{\partial \mathbf{z}_t+1}{\partial \Theta}$. 

This is how RTRL calculates its gradients without unfolding. The shape/size of $\frac{\partial \mathbf{z}_t}{\partial \Theta}$ is equal to $|z| \times |\Theta|$, therefore for standard recurrent neural networks with $n$ hidden units, this calculation scales as ${n}^4$ time complexity \cite{williams1989learning}. This makes RTRL difficult to use in practice despite having the advantage of solving the credit assignment problem. 

\textbf{Unbiased Online Recurrent Optimization:}
Unbiased Online Recurrent Optimization (UORO) \cite{tallec2017unbiased} which is considered noisy approximation of RTRL uses a rank-one trick to approximate the operations in RTRL's. This helps in reducing the overall cost during training, since the online setup UORO is faster than that for BPTT. For instance, for any given unbiased estimation of $\frac{\partial \mathbf{z}_t}{\partial \Theta}$, we can form a stochastic matrix $\tilde{Z}_t$ such that $\mathbb{E}(\tilde{Z}_{t}) = \frac{\partial \mathbf{z}_t}{\partial \Theta}$. 
Since equation \ref{rtrl:eqn2} and \ref{rtrl:eqn3} are affine in $\frac{\partial \mathbf{z}_t}{\partial \theta}$ , unbiasedness is preserved due to the linearlity of the expectation/mean. 
We compute the value of $\tilde{Z}_t$ and plug it into \ref{rtrl:eqn2} and \ref{rtrl:eqn3} to calculate the value for $\frac{\partial \mathbf{L}_t+1}{\partial \Theta}$ and $\frac{\partial \mathbf{z}_t+1}{\partial \Theta}$.
For a rank-one, unbiased approximation, we have $\tilde{Z}_t = \tilde{z}_t \otimes \tilde{\Theta}_t$ at time step $t$ .To calculate $\hat{Z}_t+1$ at $t+1$ we plug $\tilde{Z}_t$ into \ref{rtrl:eqn3}. 

In order to obtain a proper rank-one approximation, we must make use an efficient approximation technique \cite{ollivier2015training} where we rewrite the above equation as:
\begin{multline}
\tilde{Z}_{t+1} = \bigg( \rho_{0}\frac{\partial F_{\text{state}}(\mathbf{x}_{t+1},\mathbf{z}_{t},\theta)}{\partial \mathbf{z}} \tilde{\mathbf{z}}_t + \rho_{1}\nu \bigg) \otimes \bigg( \frac{\tilde{\theta}_t}{\rho_{0}} + \frac{(\nu)^T }{\rho_{1}}\frac{\partial F_{\text{state}}(\mathbf{x}_{t+1},\mathbf{z}_{t},\theta)}{\partial \theta} \bigg)
\end{multline}
where $\nu$ is a vector of independent, random signs. $\rho$ contains $k$ positive numbers and the rank one trick can be applied for any $\rho$.  In UORO, $\rho_0$ and $\rho_1$ are meant to control the variance of the derivative approximations. In practice, we define $\rho_0$ as:
\begin{align}
\rho_{0} = \sqrt[]{\frac{\| \tilde{\theta}_t \|}{\| \frac{\partial F_{state}(\mathbf{x}_{t+1},\mathbf{z}_{t},\theta)}{\partial \mathbf{z}}\tilde{\mathbf{z}}\|}} 
&\qquad
\rho_{1} = \sqrt[]{\frac{\| (\nu)^T \frac{\partial F_{state}(\mathbf{x}_{t+1},\mathbf{z}_{t},\theta)}{\theta} \|}{\|\nu \|}}\mbox{.}
\end{align}
Note that initially, $\tilde{\mathbf{z}}_0 = 0$ and $\tilde{\Theta}_0 = 0$, which yields unbiased estimates \cite{tallec2017unbiased} at time $t = 0$. Given the construction of the UORO procedure, all subsequent estimates can be shown, by induction, to be unbiased as well.

\textbf{Sparse Attentive Backtracking:}
Sparse attentive backtracking (SAB)\cite{ke2017sparse} 
is a novel approach that incorporates a differentiable, sparse attention mechanism to select from previous states.
SAB makes the following changes in a traditional network.
During forward pass, the system will manage a memory unit and also select at most a sparse subset of past memories, which is known as sparse retrieval.
During the backward pass, gradients are propagated via a sparse subset of memory and surrounding units (known as sparse replay).
Together this has shown to yield better generalization performance with improved memorization and also longer term dependency \cite{ke2017sparse}.

\textbf{Model Training Objective:}
At training time, we optimize decoder parameters with respect to a multi-objective loss over $K$-step reconstruction episodes for mini-batches of $B$ target patches $\mathbf{p}^j$ (channel input) operating over a set of decoder reconstructions $\widehat{\mathbf{p}}^j = \{ \widetilde{\mathbf{p}}^j_k, \cdots, \widetilde{\mathbf{p}}^j_K \}$ (channel outputs). The loss is defined as a convex combination of mean squared error (MSE) and mean absolute error (MAE) as follows:
\begin{align}
\mathcal{D}(\mathbf{p}^j,\widehat{\mathbf{p}}^j) = (1 - \alpha) \mathcal{D}_{MAE}(\mathbf{p}^j,\widehat{\mathbf{p}}^j) + \alpha \mathcal{D}_{MSE}(\mathbf{p}^j,\widehat{\mathbf{p}}^j) \mbox{.} \label{cost_function}
\end{align}
$\alpha$ is a tunable coefficient that controls the trade-off between the two distortion terms. In early experiments, $\alpha = 0.235$ was found to provide a good trade-off between the two (using MSE measured on the validation set as a guide).
The individual terms of the cost are:
\begin{align}
\mathcal{D}_{MAE}(\mathbf{p}^j,\widehat{\mathbf{p}}^j) = \frac{1}{(2 K)}  \sum^K_{k=1} \sum^B_{b=1} \sum_i | ( \widehat{\mathbf{p}}^{j,b}_k[i] - \mathbf{p}^{j,b}[i] ) | \label{eqn:mae} \\
\mathcal{D}_{MSE}(\mathbf{p}^j,\widehat{\mathbf{p}}^j) = \frac{1}{(2 B K)}  \sum^K_{k=1} \sum^B_{b=1} \sum_i ( \widetilde{\mathbf{p}}^{j,b}_k[i] - \mathbf{p}^{j,b}[i] )^2  \label{eqn:mse}
\end{align}
where $i$ indexes a single dimension of a vector. MSE \& MAE are used in \cite{orobiadcc} and MSE was used \cite{malidcc}.

\section{Experiments}
\label{experiments}

We implemented several variations of the recurrent state-function $\mathbf{s}_k$ for the estimator described above. In preliminary experiments, we found that the LSTM state function and the RNN-SNE (a more expensive, but expensive extension of our estimator \cite{mali2019neural}) yielded the most consistent performance. Therefore, we report the performance using an LSTM as a state cell for all algorithms and RNN-SNE using BPTT and SAB (since we found that SAB worked best when using LSTM state functions) \footnote{We conducted multiple trials on a subset of the training set and found LSTM to work stably compared to other recurrent units such as gate recurrent unit, minimal gated united, and delta RNN when trained using other learning algorithms. The second majority of these recurrent units converged to similar loss, but generalization performance varied drastically. We conducted multiple trials.}. For the SAB algorithm, we adopt the default setting proposed in \cite{ke2017sparse}.  
Beside comparing across different learning algorithm setups, we also compare our models to standard approaches including JPEG and JPEG 2000 (JP2) as well as to an end-to-end neural compression system, GOOG \cite{toderici2016full} on the image benchmarks presented in \cite{orobiadcc} (we refer the reader to this last reference for details). We conducted experiments on a subset of the training data to find optimal meta-parameters for each algorithm, since tuning these values on large scale datasets would be quite prohibitive.

\subsection{Data \& Benchmarks}  
\label{exp:data}
We adapt the setup provided in previous work \cite{orobiadcc} and create training set, that randomly samples 178k images from the \emph{Places365} {\cite{zhou2017places}} dataset, down-sampled to $512\times512$ and combine them with randomly sampled 7168 raw images from RAISE-ALL{\cite{Raise}} dataset, down-sampled to a $1600\times1600$. Training dataset were compressed using variable bit rates between $0.35$--$1.02$ bits per pixel (bpp) (once for JPEG encoder and once for JP2 encoder). Similarly, to create a validation sample, we randomly selected 20K images from the \emph{Places365} development set combined with the remaining 1K RAISE-ALL images. Validation samples were also compressed using bitrates between $0.35$--$1.02$ bpp. 
For simplicity, we focus this study on single channel images and convert each image to gray-scale. However, though we focus on gray-scale, our proposed iterative refinement can be used with other formats, e.g., RGB. We divide images into sets of $8\times8$ patches/blocks for JPEG (yielding $4096$ patches for $512\times512$ images and $40000$ for $1600\times1600$ images) and $64\times64$ patches/tiles for JP2 (producing $64$ patches for $512\times512$ images and $625$ patches for $1600\times1600$ images).

We could cut this if we think it's assumed we do this
We then further divide images into a set of $8\times8$ non-overlapping patches, which produces $4096$ patches for images of dimension $512\times512$ and $40000$ for images of size $1600$. After this we create a dataset of patch ``blocks'', or collections of nine neighboring/local patches which are fed as input to our compression model in order to predict a target patch of size $8\times8$. For example, assuming an input image is divided into $8\times8$ non overlapping patches and each block/patch is numbered sequentially starting from top-left (and ending at the bottom-right of the image), where the first index is the row number and second value is the column number, we would then use: 
$$(0,0) , (0,1), (0,2), (1,0), (1,1), (1,2), (2,0), (2,1), (2,2)$$
to predict patch $(1,1)$. Similarly, the next block would contain 
$$(0,1), (0,2), (0,3), (1,1), (1,2), (1,3),(2,1), (2,2), (2,3)$$ 
which is used to predict $(1,2)$. Similarly, in order to predict boundary blocks, such as $(0,0)$, we consider the same 9 patches as earlier, such as the set: 
$$(0,0) ,(0,1),(0,2),(1,0),(1,1),(1,2),(2,0),(2,1), (2,2)$$ 
but now our model will attempt to reconstruct $(0,0)$ instead of $(1,2)$. Doing this allows us to also reconstruct image patches at the boundary of an image using as much context as possible (9 possible input patches) without resorting to zero-padding.\footnote{We originally experimented with zero-padding, or using null context patches, when predicting boundary patches. Better reconstruction was obtained when a full block was used instead, even if some of the non-causal spatial context patches were not necessarily immediate neighbors of the target boundary patch.}

We experimented with $6$ different test sets: 1) the Kodak Lossless True Color Image Suite\footnote{\url{http://r0k.us/graphics/kodak/}} (Kodak) with 24 true color 24-bit uncompressed images, 2) the image compression benchmark (CB~8-Bit\footnote{\url{http://imagecompression.info/}}) with 14 high-resolution 8-bit grayscale uncompressed images downsampled to $1200\times1200$ images, 3) the image compression benchmark (CB~16-Bit) with 16-bit uncompressed images also downsampled to $1200\times1200$, 4) the image compression benchmark 16-bit-linear (CB~16-Bit-Linear) containing 9 high-quality 16-bit uncompressed images downsampled to $1200\times1200$, 5) Tecnick {\cite{teck}} (36 8-bit images), and 6) the Wikipedia test-set created by crawling 100 high-resolution $1200\times1200$ images from the Wikipedia website.

\setlength{\tabcolsep}{4pt}
\begin{table}[t]
\tabcolsep=1.0mm
\centering \footnotesize
\caption{\footnotesize PSNR of the LSTM-JPEG on the Kodak dataset (bitrate $0.37$ bpp) as a function of $K$. RTRL is the best at $K=1$ but all improve with extra refinement steps (SAB yields best PSNR). 
}
\label{results:kodak_progressive}
\begin{tabular}{|c||c|c|c|c|c|c|}
\hline
   & $K = 1$ & $K = 3$ & $K = 5$ & $K = 7$ & $K = 9$ & $K = 11$\\
  \hline
  BPTT &  27.0097&  27.3989 &  27.6625 &  27.8959 &  28.2199 &  \textbf{28.5099}\\
  SAB & {27.1009} & \textbf{27.9962} & \textbf{28.853} & \textbf{28.844} & \textbf{28.7911} & {28.4566} \\
  UORO & 27.1001& 27.4411& 27.5589& 27.9912& 28.001& 27.999\\
  RTRL & \textbf{27.20001}& 27.3555& 27.4888& 27.8888& 28.01188& 27.0012\\
  \hline
\end{tabular}
\begin{center}
\label{results:samples1}
\caption{PSNR for iterative decoding using BPTT.}\vspace{-0.2cm} 
\footnotesize
\begin{tabular}{c c c c}
\hline
\includegraphics[width=0.2\textwidth]{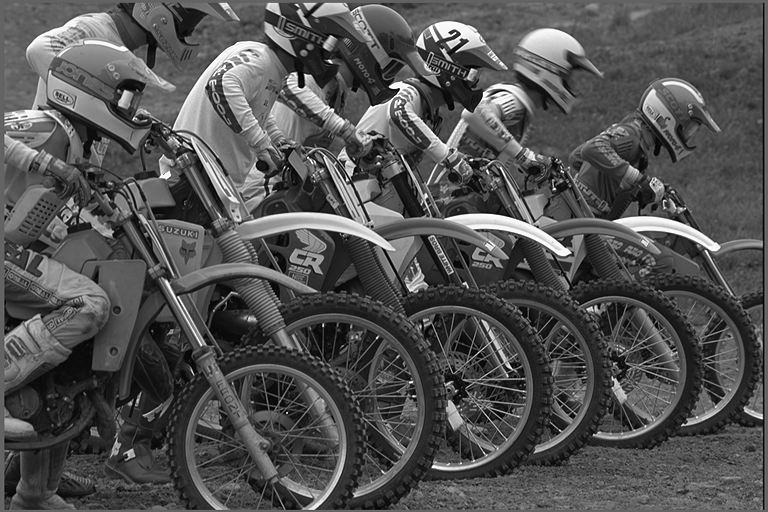} & \includegraphics[width=0.2\textwidth]{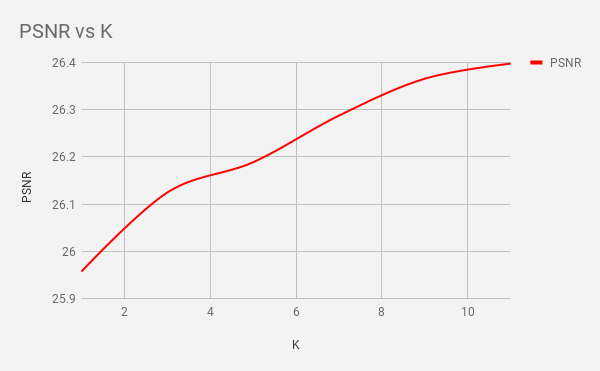} 
\includegraphics[width=0.2\textwidth]{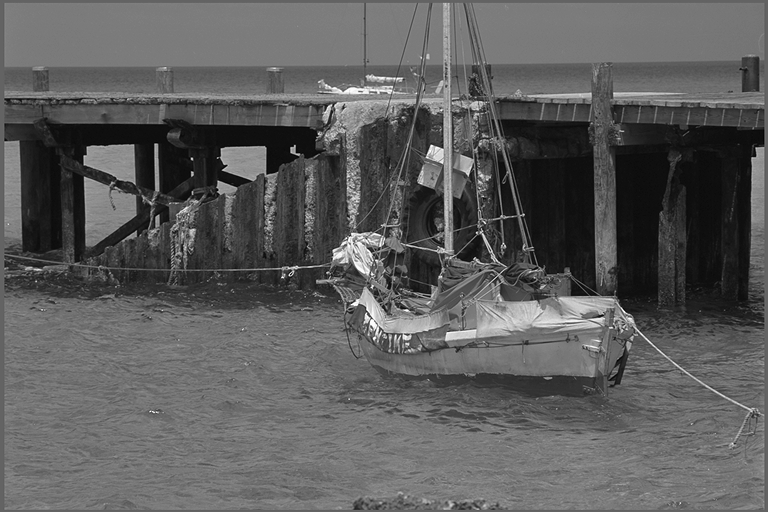}  & \includegraphics[width=0.2\textwidth]{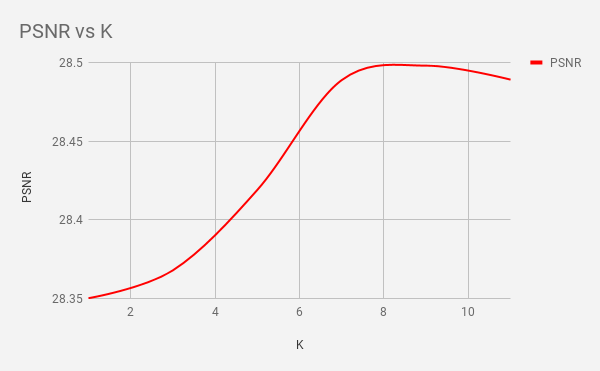} \\
\hline
\includegraphics[width=0.2\textwidth]{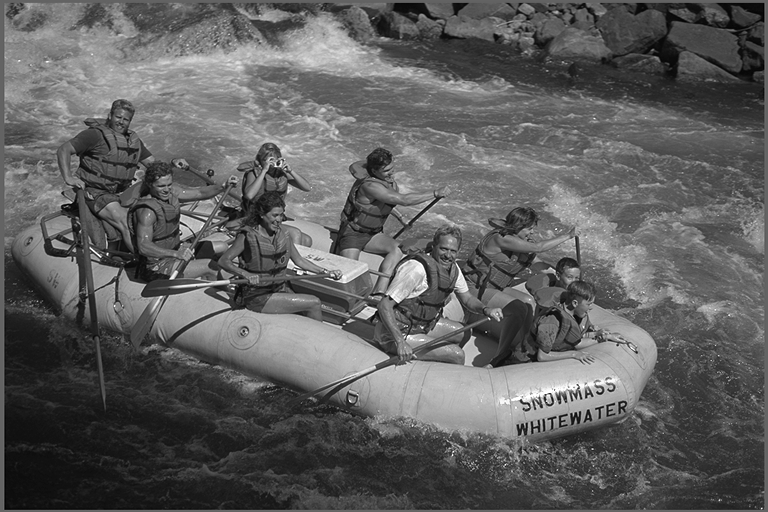}  & \includegraphics[width=0.2\textwidth]{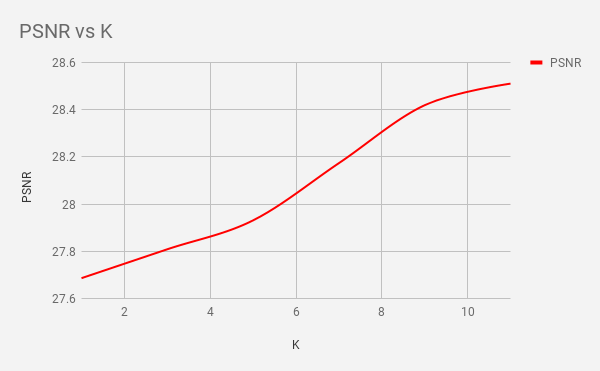} 
\includegraphics[width=0.2\textwidth]{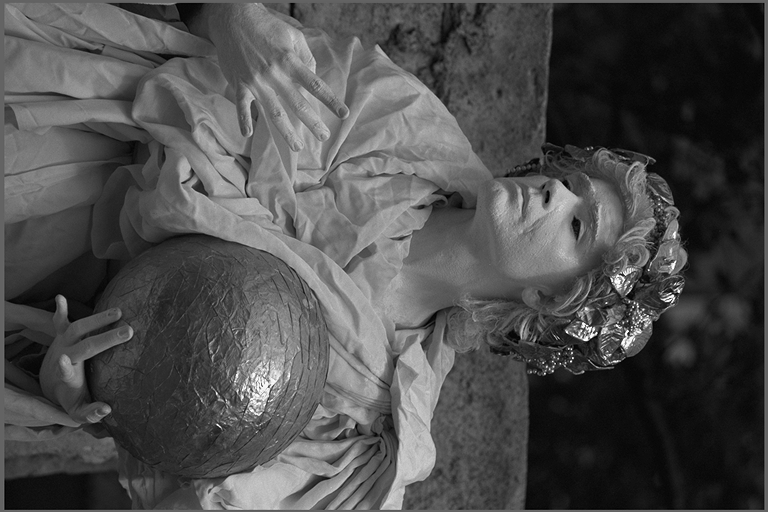}  & \includegraphics[width=0.2\textwidth]{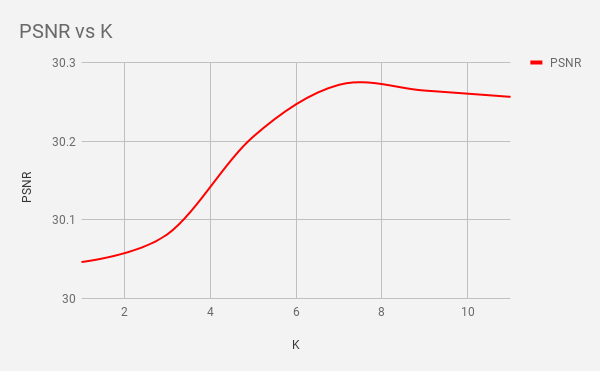} \\
\end{tabular}
\end{center}
\end{table}

\subsection{Experimental Setup}
\label{exp:design}
All of our recurrent estimators consist of one hidden layer with $512$ units. We initialize weights from a uniform distribution, $\sim U(-0.054, 0.054)$. Using the updates from each algorithm, model weights were updated via stochastic gradient descent (SGD) with mini-batches of $256$ and L2 polynomial decay -- initial step size was $2e-4$. Gradients norms were clipped to $13$ and models were trained for more than $260$ epochs. We also used the two-step data shuffling technique of \cite{orobiadcc} and we set $K = 4$ (though better convergence was observed for SAB and UORO with $K = 3$). Dataset and experimental settings are identical to prior work \cite{orobiadcc, malidcc}.


\subsection{Results}
We evaluated our model on $6$ benchmarks used in prior work \cite{orobiadcc} using $3$ metrics \cite{ma2016group}. These metrics are Peak Signal to Noise Ratio (PSNR), structural similarity (SSIM), and multi-scale structural similarity (MS-SSIM \cite{wang2004imagequality}, or $MS^3IM$.

 We report results in Table \ref{results:benchmarks} for BPTT, SAB, UORO, RTRL on the Kodak dataset. All learning approaches can be seen consistently yielding lower distortion reconstruction as compared to JPEG, JP2, GOOG\cite{toderici2016full}, and E2E \cite{balle2016end}. Worthy of note, when using SAB, we noticed that there was a tendency to memorize the previous patch pattern for a longer duration compared to BPTT and other approaches \cite{ke2017sparse}.
In terms of PSNR (on Kodak), we achieve nearly a $1.6049$ decibel (dB) gain (with \emph{LSTM}-JP2-SAB) over JPEG and a $1.4219$ dB gain (with \emph{LSTM}-JP2-SAB) over JP2 when using SAB. With respect to GOOG, our \emph{LSTM}-JP2-SAB estimator yields a gain of $1.2976$ dB and when compared with \emph{LSTM}-JP2-BPTT, our \emph{LSTM}-JP2-SAB achieves a $0.3277$ dB improvement. The results, across all benchmark test-sets, for all metrics (PSNR, SSIM, and MS-SSIM), show that decoders learned with SAB and iterative refinement generate images with lower distortion and higher perceptual quality (as indicated by SSIM \& MS-SSIM). One observation is that, according to our experiments, UORO and RTRL, despite having low PSNRs, have reasonably good perceptual quality (SSIM and $MS^3IM$ -- these two values are the same for both algorithms) as seen in Table \ref{results:benchmarks}). In addition, note that, besides BPTT, all approaches require smaller $K$. 

In Table \ref{results:kodak_progressive}, we present our best-performing LSTM and show how PSNR varies as a function of $K$ (number of iterative refinement steps). We analyze this performance for all learning algorithms and noticed that SAB consistently outperforms the rest. Another finding based on our analysis is that, for $K=1$, all methods yield better PSNR's better than BPTT, but a few of them degrade as $K$ increases. We also see that raising $K$ usually improves image reconstruction with respect to PSNR. In Table 2,
\ref{results:samples1} we sampled four random images and plotted their PSNR as a function of $K$ for BPTT. In general, increasing $K$ seems to improve PSNR, but, in some cases, we see diminishing returns.

\begin{table}[!t]\renewcommand{\arraystretch}{1.0}
\begin{center}
\caption{Out-of-sample results for the Kodak (bpp $0.37$), 8-bit Compression Benchmark (CB, bpp, $0.341$), 16-bit \& 16-bit-Linear Compression Benchmark (CB) datasets (bpp $0.35$ for both), the Tecnick (bpp $0.475$), \& Wikipedia (bpp $0.352$) datasets.}
\label{results:benchmarks}

\resizebox{\textwidth}{!}{%
\begin{tabular}{|l||c|c|c|c||c|c|c|c|}
\hline
  & \multicolumn{3}{c||}{\textbf{Kodak}} & \multicolumn{3}{c|}{\textbf{CB 8-Bit}}\\
  \textbf{Model} & \textbf{PSNR} &  \textbf{SSIM} &  $MS^3IM$ & \textbf{PSNR} & \textbf{SSIM} & $MS^3IM$\\
  \hline
  \emph{JPEG} & 27.6540 & 0.7733 & 0.9291 & 27.5481 &  0.8330 & 0.9383 \\
  \emph{JPEG 2000} & 27.8370  & 0.8396 & 0.9440 & 27.7965 &  0.8362 & 0.9471 \\
  \emph{GOOG}-JPEG & 27.9613  & 0.8017 & 0.9557 & 27.8458 &  0.8396 & 0.9562 \\
  \emph{E2E} (\textbf{Neural}) & 28.9420  & 0.8502 & 0.9600 & 28.0999  & 0.8396 & 0.9562 \\
 \emph{LSTM}-JP2 - BPTT & {28.9321}  & {0.8425} & {0.9596} & {28.0896}  & {0.8389} & {0.9562} \\ 
 \emph{LSTM}-JP2 - SAB & {29.2589} & {0.8435} & {0.9599} & {28.2896} & {0.8399} & {0.9577} \\ 
 \emph{LSTM}-JP2 - UORO & {28.9302}  & {0.8424} & {0.9592} & {28.0895} & {0.8388} & {0.9562} \\ 
 \emph{LSTM}-JP2 - RTRL & {28.9311} & {0.8424} & {0.9595} & {28.0891}  & {0.8387} & {0.9561} \\ 
 \emph{SNE-RNN}-JP2 - BPTT & {29.3008}  & {0.8508} & {0.9622} & {28.2199} & {0.8401} & {0.9600} \\
 \emph{SNE-RNN}-JP2 - SAB & \textbf{29.4128}  & \textbf{0.8514} & \textbf{0.9627} & \textbf{28.2122} & \textbf{0.8402} & \textbf{0.9600} \\
  
  \hline
    & \multicolumn{3}{c||}{\textbf{CB 16-Bit}} & \multicolumn{3}{c|}{\textbf{CB 16-Bit-Linear}}\\
  \hline
  \emph{JPEG} &27.5368  & 0.8331 & 0.9383 & 31.7522  & 0.8355 & 0.9455 \\
  \emph{JPEG 2000} & 27.7885  & 0.8391 & 0.9437 & 32.0270 &  0.8357 & 0.9471 \\
  \emph{GOOG} & 27.8830 & 0.8391 & 0.9468  & 32.1275 & 0.8369 & 0.9533 \\
   \emph{E2E} (\textbf{Neural}) & 28.2440 & 0.8426 & 0.9498  & 32.5010 & 0.8387 & 0.9540 \\
  \emph{LSTM}-JP2- BPTT & {28.1307} & {0.8425} & {0.9496} & {32.4998}  & {0.8382} & {0.9541} \\
   \emph{LSTM}-JP2 - SAB & {28.2307} & {0.8431} & {0.9501} & {32.5003} & {0.8385} & {0.9544} \\
    \emph{LSTM}-JP2 - UORO & {28.1312} & {0.8427} & {0.9497} & {32.4999}  & {0.8382} & {0.9542} \\
     \emph{LSTM}-JP2 - RTRL & {28.1304} & {0.8424} & {0.9495} & {32.4991} & {0.8381} & {0.9542} \\
     \emph{SNE-RNN}-JP2- BPTT & {29.4471} &  {0.8430} & {0.9510} & \textbf{32.6019} & \textbf{0.8399} & {0.9559} \\
      \emph{SNE-RNN}-JP2- SAB & \textbf{29.4524} &  \textbf{0.8432} & \textbf{0.9511} & {32.6011} & {0.8398} & \textbf{0.9560} \\
  
  \hline
  & \multicolumn{3}{c||}{\textbf{Tecnick}} & \multicolumn{3}{c|}{\textbf{Wikipedia}}\\
  \hline
  \emph{JPEG} &  30.7377  & 0.8682 &  0.9521 & 28.7724  & 0.8290 &  0.9435\\
  \emph{JPEG 2000} & 31.2319  & 0.8747 & 0.9569 & 29.1545 &  0.8382 & 0.9495 \\
  \emph{GOOG} & 31.5030& 0.8814 & 0.9608 & 29.2209   &  0.8406 &  0.9520 \\
  \emph{E2E} (\textbf{Neural}) & 31.7000  & 0.8836 & 0.9620 & 29.3227  &  0.8412 &  0.9526 \\
  \emph{LSTM}-JP2 - BPTT & {31.6962}  & {0.8834} & {0.9619} & {29.3228} & {0.8411} & {0.9526} \\
  \emph{LSTM}-JP2 - SAB & {31.7100}  & {0.8841} & {0.9621} & {29.3229}& {0.8412} & {0.9525} \\
  \emph{LSTM}-JP2 - UORO & {31.6961} & {0.8835} & {0.9620} & {29.3225}  & {0.8412} & {0.9526} \\
  \emph{LSTM}-JP2- RTRL & {31.6963}  & {0.8833} & {0.9615} & {29.3227} & {0.8412} & {0.9525} \\
  \emph{SNE-RNN}-JP2 - BPTT  & \textbf{32.7124} &  {0.8841} & \textbf{0.9622} & {29.9334} &  {0.8413} & {0.9528} \\
  \emph{SNE-RNN}-JP2 - SAB  & {32.7119} &  \textbf{0.8844} & {0.9621} & \textbf{29.9336} &  \textbf{0.8414} & \textbf{0.9529} \\
  \hline
\end{tabular}
}
\end{center}
\end{table}


\section{Discussion}
\label{sec:discussion}
Prior work on image compression has focused on optimizing models to achieve better reconstruction at lower bit rates. These approaches are focused on creating sophisticated architectures that enhance encoder or decoder performance. In some cases, there is the desire to jointly optimize both along with a designed form of entropy encoding. In some instances, these approaches result in the creation of many redundant components, which may or may not be useful. Knowing that BP-based learning is plagued with a variety of credit assignment-related issues, adding more components could make performance even worse. Additionally, adding more components to the system means adding more trainable parameters which complicates the training process and resulting in sub-optimal performance. 

According to our experiments, SAB and UORO learning algorithms seem to be the most promising alternatives. Though SAB is still essentially BPTT-based, it combines memory replay with a a sparse memory retrieval scheme that reduces some computational burden while maintaining state memory for longer time spans. Another observation of our work is that memorization (or model statefulness) is necessary for capturing the longer-term dependencies implicit to the act of compressing iteratively (even though non-temporal inputs are inputs). A more powerful model with memory can also help in reaching better PSNR when trained across a global set of patches. Note that algorithms like RTRL and UORO (desirably) do not require unrolling the estimator over $K$ steps in time, reducing sequence data storage requirements. However, the noisy rank one approximation trick used in UORO affects the memorization pattern and seems to prevent it from reaching the absolute best performance. Despite such drawbacks, UORO generates good reconstruction (w.r.t. perceptual quality). 

In our experiments, we learned that online, forward differentiation methods, i.e., UORO and RTRL, struggle with adaptive learning rate rules, e.g., Adam \& Adagrad (we found that SGD and simple momentum are best used with these procedures), resulting in less of a performance gain than with BPTT and SAB. Furthermore, mechanisms such dropout, zone-out, and batch normalization appear to hurt the performance of UORO and RTRL. Unfortunately, this appears to imply that backprop-based heuristics do not easily/readily work favorably with online, forward differentiation methods. We also found that UORO was quite sensitive to hidden layer sizes and interacted negatively more often with various optimizers as compared to RTRL. As a result, one needs to perform a careful grid search to obtain best settings. SAB was also found to be sensitive to hyper-parameters and, in our experiments, dropout on non-recurrent connections hurt its  performance considerably (compared to BPTT). Notably, of its specific meta-parameters, i.e., $k_{top}$ and $k_{attn}$, SAB appears to be much more sensitive to the choice of $k_{top}$ as compared to $k_{attn}$. In the end, all algorithms seem to yield different solutions though seem to overall yield better decoders. This work represents an important step towards re-imagining the design of neural-based compression systems, showing specifically that by changing how the parameters are adjusted, we might find yet further gains in system performance.

\begin{figure}[h]
\centering
\includegraphics[width=0.625\textwidth]{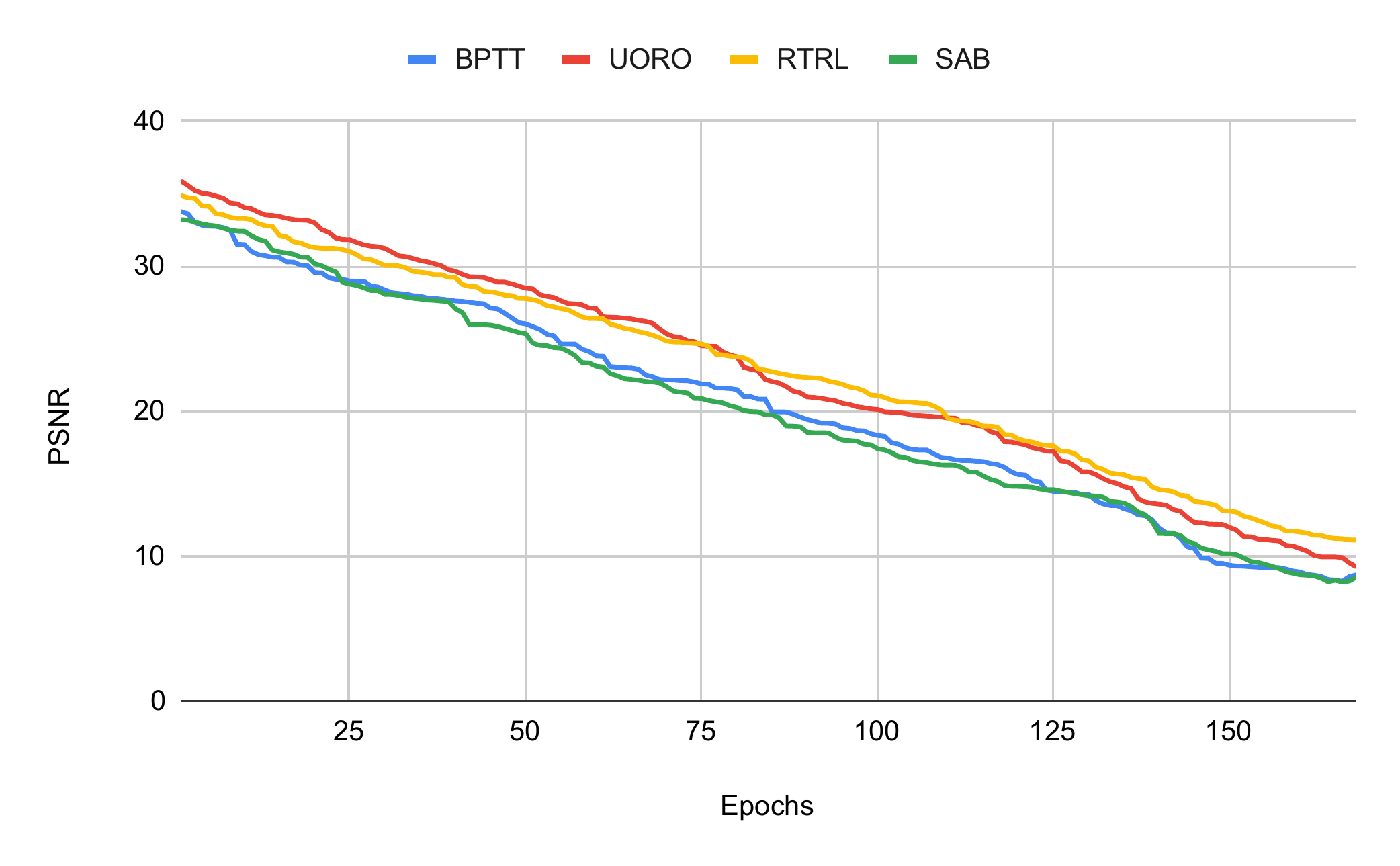}
\vspace{-0.3cm}
\caption{PSNR as a function of epoch (measurements made on validation set). 
}
\label{fig:psnrcurve}
\end{figure}

\noindent
\textbf{Convergence:} We plot, in Figure \ref{fig:psnrcurve}, the validation PSNR curves of our best RNN deocder trained with different algorithms.
Notice that SAB, which improves RNN memory retention, converges slightly better/sooner than BPTT.

\section{Conclusions}
\label{sec:conc}
We analyzed the effect of various popular recurrent learning algorithms using \emph{iterative refinement} on hybrid compression systems for lossy image compression. Interestingly, variants of BPTT yield better performance across several metrics, seemingly in some cases offering an improved ability to improve memory retention in RNN decoder estimators when attempting to iteratively compress an image composed of many blocks/patches. We trained a neural decoder using several different recurrent learning algorithms and compared the resultant models to  standard JPEG, JPEG-2000, and two competitive neural compression systems, GOOG \& E2E. The hybrid decoder trained using sparse attentive backtracking (SAB) performed the best among all approaches at low bit rates. Surprisingly, SAB reduced the number of iterative steps needed to compress a complex image compared to other approaches. 
Future work could focus on analyzing perceptual quality at lower bit rates for these recurrent learning algorithms for more complex, end-to-end neural compression systems, potentially uncovering other benefits and limitations of these procedures.;
Other work could entail building principled components inspired by signal processing that could reduce additional (design) redundancy in end-to-end systems while interacting favorably with online recurrent learning approaches.  

\Section{References}
\bibliographystyle{IEEEbib}
\bibliography{zmain}

\begin{thebibliography}{10}

\bibitem{orobiadcc}
A.~G. {Ororbia}, A.~{Mali}, J.~{Wu}, S.~{O'Connell}, W.~{Dreese}, D.~{Miller},
  and C.~L. {Giles},
\newblock ``Learned neural iterative decoding for lossy image compression
  systems,''
\newblock in {\em DCC}, March 2019, pp. 3--12.

\bibitem{malidcc}
A.~{Mali}, A.~G. {Ororbia}, and C.~L. {Giles},
\newblock ``The sibling neural estimator: Improving iterative image decoding
  with gradient communication,''
\newblock in {\em DCC}, 2020, pp. 23--32.

\bibitem{toderici2016full}
George Toderici, Damien Vincent, Nick Johnston, Sung~Jin Hwang, David Minnen,
  Joel Shor, and Michele Covell,
\newblock ``Full resolution image compression with recurrent neural networks,''
\newblock {\em CoRR}, vol. abs/1608.05148, 2016.

\bibitem{balle2016end}
Johannes Ball{\'e}, Valero Laparra, and Eero~P Simoncelli,
\newblock ``End-to-end optimized image compression,''
\newblock {\em arXiv preprint arXiv:1611.01704}, 2016.

\bibitem{williams1989experimental}
Ronald~J Williams and David Zipser,
\newblock ``Experimental analysis of the real-time recurrent learning
  algorithm,''
\newblock {\em Connection Science}, vol. 1, no. 1, pp. 87--111, 1989.

\bibitem{tallec2017unbiased}
Corentin Tallec and Yann Ollivier,
\newblock ``Unbiased online recurrent optimization,'' 2017.

\bibitem{ke2017sparse}
Nan~Rosemary Ke, A.~G., Olexa Bilaniuk, Jonathan Binas, Laurent Charlin, Chris
  Pal, and Yoshua Bengio,
\newblock ``Sparse attentive backtracking: Long-range credit assignment in
  recurrent networks,''
\newblock {\em arXiv preprint arXiv:1711.02326}, 2017.

\bibitem{takamura1994coding}
S.~Takamura and M.~Takagi,
\newblock ``Lossless image compression with lossy image using adaptive
  prediction and arithmetic coding,''
\newblock in {\em DCC}, March 1994, pp. 166--174.

\bibitem{oord2016pixel}
A{\"{a}}ron van~den Oord, Nal Kalchbrenner, and Koray Kavukcuoglu,
\newblock ``Pixel recurrent neural networks,''
\newblock in {\em ICML}, 2016, pp. 1747--1756.

\bibitem{gregor2016conceptualcompression}
Karol Gregor, Frederic Besse, Danilo Jimenez~Rezende, Ivo Danihelka, and Daan
  Wierstra,
\newblock ``Towards conceptual compression,''
\newblock in {\em NIPS}, pp. 3549--3557. 2016.

\bibitem{toderici2015}
George Toderici, Sean~M. O'Malley, Sung~Jin Hwang, Damien Vincent, David
  Minnen, Shumeet Baluja, Michele Covell, and Rahul Sukthankar,
\newblock ``Variable rate image compression with recurrent neural networks,''
\newblock {\em CoRR}, vol. abs/1511.06085, 2015.

\bibitem{johnston2018improved}
Nick Johnston, Damien Vincent, David Minnen, Michele Covell, Saurabh Singh,
  Troy Chinen, Sung Jin~Hwang, Joel Shor, and George Toderici,
\newblock ``Improved lossy image compression with priming and spatially
  adaptive bit rates for recurrent networks,''
\newblock in {\em CVPR}, 2018, pp. 4385--4393.

\bibitem{Theis2017}
Lucas Theis, Wenzhe Shi, Andrew Cunningham, and Ferenc Husz{\'{a}}r,
\newblock ``Lossy image compression with compressive autoencoders,''
\newblock {\em CoRR}, vol. abs/1703.00395, 2017.

\bibitem{rippel2017real}
Oren Rippel and Lubomir~D. Bourdev,
\newblock ``Real-time adaptive image compression,''
\newblock in {\em ICML}, 2017, pp. 2922--2930.

\bibitem{balle2018variational}
Johannes Ball{\'e}, David Minnen, Saurabh Singh, Sung~Jin Hwang, and Nick
  Johnston,
\newblock ``Variational image compression with a scale hyperprior,''
\newblock {\em arXiv preprint arXiv:1802.01436}, 2018.

\bibitem{agustsson2017soft}
Eirikur Agustsson, Fabian Mentzer, Michael Tschannen, Lukas Cavigelli, Radu
  Timofte, Luca Benini, and Luc~V Gool,
\newblock ``Soft-to-hard vector quantization for end-to-end learning
  compressible representations,''
\newblock in {\em NIPS}, 2017, pp. 1141--1151.

\bibitem{cheng2019learning}
Zhengxue Cheng, Heming Sun, Masaru Takeuchi, and Jiro Katto,
\newblock ``Learning image and video compression through spatial-temporal
  energy compaction,''
\newblock in {\em CVPR}, 2019.

\bibitem{chen19}
Z.~{Cheng}, H.~{Sun}, M.~{Takeuchi}, and J.~{Katto},
\newblock ``Energy compaction-based image compression using convolutional
  autoencoder,''
\newblock {\em IEEE Transactions on Multimedia}, pp. 1--1, 2019.

\bibitem{dcc19}
S.~{Li}, Z.~{Zheng}, W.~{Dai}, and H.~{Xiong},
\newblock ``Lossy image compression with filter bank based convolutional
  networks,''
\newblock in {\em DCC}, March 2019, pp. 23--32.

\bibitem{strumpler2020learning}
Yannick Str{\"u}mpler, Ren Yang, and Radu Timofte,
\newblock ``Learning to improve image compression without changing the standard
  decoder,''
\newblock {\em arXiv preprint arXiv:2009.12927}, 2020.

\bibitem{lee2015difference}
Dong-Hyun Lee, Saizheng Zhang, Asja Fischer, and Yoshua Bengio,
\newblock ``Difference target propagation,''
\newblock in {\em ECML PKDD}. Springer, 2015, pp. 498--515.

\bibitem{nokland2016direct}
Arild N{\o}kland,
\newblock ``Direct feedback alignment provides learning in deep neural
  networks,''
\newblock in {\em NIPS}, 2016, pp. 1037--1045.

\bibitem{ororbia2019biologically}
Alexander~G Ororbia and Ankur Mali,
\newblock ``Biologically motivated algorithms for propagating local target
  representations,''
\newblock in {\em AAAI}, 2019, vol.~33, pp. 4651--4658.

\bibitem{ororbia2018continual}
Alexander Ororbia, Ankur Mali, C~Lee Giles, and Daniel Kifer,
\newblock ``Continual learning of recurrent neural architectures by locally
  aligning distributed representations,''
\newblock {\em arXiv preprint arXiv:1810.07411}, 2018.

\bibitem{werbos1988generalization}
Paul~J Werbos,
\newblock ``Generalization of backpropagation with application to a recurrent
  gas market model,''
\newblock {\em Neural networks}, vol. 1, no. 4, pp. 339--356, 1988.

\bibitem{werbos1990backpropagation}
Paul~J Werbos,
\newblock ``Backpropagation through time: what it does and how to do it,''
\newblock {\em Proceedings of the IEEE}, vol. 78, no. 10, pp. 1550--1560, 1990.

\bibitem{mali2019neural}
Ankur Mali, Alexander Ororbia, and C.~Lee Giles,
\newblock ``The neural state pushdown automata,'' 2019.

\bibitem{ororbia2017diff}
Alexander~G. Ororbia, II, Tomas Mikolov, and David Reitter,
\newblock ``Learning simpler language models with the differential state
  framework,''
\newblock {\em Neural Comput.}, vol. 29, no. 12, pp. 3327--3352, Dec. 2017.

\bibitem{hochreiter1997long}
Sepp Hochreiter and J{\"u}rgen Schmidhuber,
\newblock ``Long short-term memory,''
\newblock {\em Neural Computation}, vol. 9, no. 8, pp. 1735--1780, 1997.

\bibitem{chung2014empirical}
Junyoung Chung, Caglar Gulcehre, KyungHyun Cho, and Yoshua Bengio,
\newblock ``Empirical evaluation of gated recurrent neural networks on sequence
  modeling,''
\newblock {\em arXiv preprint arXiv:1412.3555}, 2014.

\bibitem{williams1989learning}
Ronald~J Williams and David Zipser,
\newblock ``A learning algorithm for continually running fully recurrent neural
  networks,''
\newblock {\em Neural computation}, vol. 1, no. 2, pp. 270--280, 1989.

\bibitem{ollivier2015training}
Yann Ollivier, Corentin Tallec, and Guillaume Charpiat,
\newblock ``Training recurrent networks online without backtracking,''
\newblock {\em arXiv preprint arXiv:1507.07680}, 2015.

\bibitem{zhou2017places}
Bolei Zhou, {\`{A}}gata Lapedriza, Aditya Khosla, Aude Oliva, and Antonio
  Torralba,
\newblock ``Places: {A} 10 million image database for scene recognition,''
\newblock {\em {IEEE} Trans. Pattern Anal. Mach. Intell.}, vol. 40, no. 6, pp.
  1452--1464, 2018.

\bibitem{Raise}
Duc-Tien Dang-Nguyen, Cecilia Pasquini, Valentina Conotter, and Giulia Boato,
\newblock ``Raise: A raw images dataset for digital image forensics,''
\newblock in {\em Proceedings of the 6th ACM Multimedia Systems Conference},
  New York, NY, USA, 2015, MMSys '15, pp. 219--224, ACM.

\bibitem{teck}
Nicola Asuni and Andrea Giachetti,
\newblock ``Testimages: A large data archive for display and algorithm
  testing,''
\newblock {\em Journal of Graphics Tools}, vol. 17, no. 4, pp. 113--125, 2013.

\bibitem{ma2016group}
Kede Ma, Qingbo Wu, Zhou Wang, Zhengfang Duanmu, Hongwei Yong, Hongliang Li,
  and Lei Zhang,
\newblock ``Group {MAD} competition? {A} new methodology to compare objective
  image quality models,''
\newblock in {\em CVPR}, 2016, pp. 1664--1673.

\bibitem{wang2004imagequality}
Zhou Wang, Alan~C. Bovik, Hamid~R. Sheikh, and Eero~P. Simoncelli,
\newblock ``Image quality assessment: from error visibility to structural
  similarity,''
\newblock {\em {IEEE} Trans. Image Processing}, vol. 13, no. 4, pp. 600--612,
  2004.

\end{thebibliography}

\end{document}